\documentclass[letterpaper, 10pt, reqno]{amsart}
\usepackage{mathptmx}
\pdfpageattr {/Group << /S /Transparency /I true /CS /DeviceRGB>>}

\usepackage[usenames,dvipsnames,svgnames,table]{xcolor}
\usepackage{amsmath, amsthm, amssymb, bbm, bm, mathtools}
\usepackage{enumerate}
\usepackage{graphicx}
\usepackage{float}
\usepackage{wrapfig, subcaption}
\usepackage[margin=0cm]{caption}
\usepackage[titletoc, toc]{appendix}
\usepackage{tabularx}
\usepackage{booktabs,colortbl}
\usepackage{nicefrac}
\usepackage{blindtext}
\usepackage{setspace}
\usepackage{marginnote}
\usepackage{multirow}

\usepackage[boxruled, vlined, linesnumbered]{algorithm2e}
\SetAlFnt{\small}
\SetAlCapFnt{\small}
\SetAlCapNameFnt{\small}
\usepackage{algorithmic}
\algsetup{linenosize=\tiny}

\let\oldnl\nl
\newcommand{\nonl}{\renewcommand{\nl}{\let\nl\oldnl}}

\usepackage[bookmarks=false]{hyperref}
\hypersetup{
colorlinks=true, linkcolor=Sepia, citecolor=Sepia, filecolor=magenta, urlcolor=NavyBlue,
}
\usepackage{url, bookmark}

\usepackage[capitalize,nameinlink]{cleveref}
\crefformat{equation}{(#2#1#3)}
\crefformat{theorem}{Theorem ~#2#1#3}
\crefformat{lemma}{Lemma~#2#1#3}
\crefformat{remark}{Remark~#2#1#3}
\crefformat{section}{Section~#2#1#3}
\crefrangeformat{section}{Sections~#3#1#4--#5#2#6}
\crefrangeformat{equation}{~(#3#1#4--#5#2#6)}
\crefrangeformat{figure}{Figs.~#3#1#4--#5#2#6}

\newcommand{\reals}{\mathbb{R}}

\newcommand{\norm}[1]{\ensuremath \lVert#1\rVert}

\newcommand{\ag}[1]{\ensuremath \left\langle#1\right\rangle}
\providecommand{\OO}{\mathcal{O}}

\DeclareMathOperator*{\argmin}{arg\;min}

\newcommand{\aeq}[1]{\begin{align} #1 \end{align}}

\newcommand{\beq}[1]{\begin{equation}#1\end{equation}}

\newcommand{\trm}[1]{\mathrm{#1}}

\providecommand\f[2]{\ensuremath \frac{#1}{#2}}
\providecommand\rbrac[1]{\ensuremath \left(#1\right)}
\providecommand\sqbrac[1]{\ensuremath \left[#1\right]}

\DeclarePairedDelimiter{\floor}{\lfloor}{\rfloor}

\newcommand{\grad}{\nabla}

\newtheorem{theorem}{Theorem}

\theoremstyle{definition}

\newtheorem{remark}[theorem]{Remark}

\definecolor{color_skyblue}{rgb}{0.01,0.39,0.75}

\renewcommand{\r}{\rho}

\renewcommand{\a}{\alpha}

\newcommand{\g}{\gamma}
\renewcommand{\d}{\delta}

\def \OO {\mathcal{O}}

\newcommand{\lenet}{\trm{LeNet}}
\newcommand{\allcnn}{\textrm{All-CNN}}

\newcommand{\wrncifar}{\textrm{WRN-28-10}}
\newcommand{\wrnsvhn}{\textrm{WRN-16-4}}
\newcommand{\wrnsvhnl}{\textrm{WRN-16-8}}

\newcommand{\parle}{\textrm{Parle}}

\usepackage{natbib}

\usepackage[margin=1.25in]{geometry}
\graphicspath{{./fig/}}

\newcommand{\ignore}[1]{}

\title{Parle: parallelizing stochastic gradient descent}

\author[Chaudhari, Baldassi, Zecchina, Soatto, Talwalkar and Oberman]{
\footnotesize Pratik Chaudhari$^1$, Carlo Baldassi$^2$, Riccardo Zecchina$^2$, Stefano Soatto$^1$,\\ Ameet Talwalkar$^1$ and Adam Oberman$^3$}

\begin{document}
\maketitle
{
\vspace*{-0.15in}
\footnotesize
\noindent $^{1}$ Computer Science Department, University of California, Los Angeles.\\
$^{2}$ Institute for Data Science and Analytics, Bocconi University, Milano.\\
$^{3}$ Department of Mathematics and Statistics, McGill University, Montreal.\\[0.05in]
Email:\ \href{mailto:pratikac@ucla.edu}{pratikac@ucla.edu},
\href{mailto:carlo.baldassi@unibocconi.it}{carlo.baldassi@unibocconi.it},
\href{mailto:riccardo.zecchina@unibocconi.it}{riccardo.zecchina@unibocconi.it},
\href{mailto:soatto@ucla.edu}{soatto@ucla.edu},\\
\hspace*{0.32in}\href{mailto:ameet@cs.ucla.edu}{ameet@cs.ucla.edu},
\href{mailto:adam.oberman@mcgill.ca}{adam.oberman@mcgill.ca}}\\

{
\small
\noindent \textbf{Abstract:}
We propose a new algorithm called $\parle$ for parallel training of deep networks that converges $2$-$4\times$ faster than a data-parallel implementation of SGD, while achieving significantly improved error rates that are nearly state-of-the-art on several benchmarks including CIFAR-10 and CIFAR-100, without introducing any additional hyper-parameters. We exploit the phenomenon of flat minima that has been shown to lead to improved generalization error for deep networks. Parle requires very infrequent communication with the parameter server and instead performs more computation on each client, which makes it well-suited to both single-machine, multi-GPU settings and distributed implementations.\\[-0.1in]
}

{\small \noindent \textbf{Keywords:} parallelization of SGD, deep neural networks, non-convex optimization, local entropy, robust ensembles\\[-0.1in]}

\section{Introduction}
\label{s:intro}

The dramatic success of deep networks has fueled the growth of massive datasets, e.g. Google's JFT dataset has 100 million images, this in turn has prompted researchers to employ even larger models. Parallel and distributed training of deep learning is paramount to tackle problems at this scale. Such an escalation however hits a roadblock: to minimize communication costs, one could use large batch sizes in stochastic gradient descent (SGD) but this leads to a degradation of the generalization performance.\footnote{We discuss the connection to the recent work of~\citet{goyal2017accurate} in~\cref{ss:related work}} On the other hand, small batches incur communication costs that quickly dwarf the benefits of parallelization.

In this paper, we take a different approach than model or data parallelization. The algorithm we introduce, called $\parle$, trains multiple copies of the same model in parallel. Each of these copies performs multiple gradient steps and communicates its progress very infrequently to the master parameter server. This approach is motivated by the phenomenon of ``flat minima'' that has been show to improve generalization performance of deep networks. $\parle$ has few low communication requirements and is thus well suited to both single-machine multi-GPU settings and distributed implementations over multiple compute nodes. We demonstrate extensive empirical evidence that it obtains significant performance improvements over baseline models and obtains nearly state-of-the-art generalization errors; it also obtains a $2$-$4\times$ wall-clock time speedup over data-parallel SGD. Moreover, $\parle$ is insensitive to hyper-parameters, all experiments in this paper are conducted with the same hyper-parameters. We do not introduce any additional hyper-parameters over SGD.

\subsection{Approach}
\label{ss:approach}

If we denote the parameters of a deep network by $x \in \reals^N$, training consists of solving
\beq{
    x^* = \argmin_x\ f(x),
    \label{prob:org}
}
where $f(x)$ is the average loss (say cross-entropy) over the entire dataset, along with a regularization term (say weight decay). We denote with $x^1, \ldots, x^n$ copies of the model by variables, also called ``replicas.'' They may reside on multiple GPUs on the same computer, on multiple compute nodes. As a means of coupling these variables, consider the loss function of Elastic-SGD~\citep{zhang2015deep}:
\beq{
    \argmin_{x,\ x^1,\ \ldots,\ x^n}\ \sum_{a=1}^n\ f(x^a) + \f{1}{2 \r}\ \norm{x^a - x}^2;
    \label{prob:esgd}
}
where a parameter $\r > 0$ couples two replicas $x^a$ and $x^b$ through a ``reference'' variable $x$. Performing gradient descent on~\cref{prob:esgd} involves communicating the replicas $x^a$ for all $a \leq n$ with the reference after each mini-batch. Even though Elastic-SGD was introduced in the parallel setting, it nonetheless introduces significant communication bottlenecks.

In order to reduce this communication, we replace $f(x)$ by a modified loss function called ``local entropy''
\beq{
    f_\g(x) := -\log \rbrac{G_\g\ *\ e^{-f(x)}};
    \label{eq:le}
}
where
\[
    G_\g = (2 \pi \g)^{-N/2}\ \exp \rbrac{-\f{\norm{x}^2}{2 \g}}
\]
is the Gaussian kernel with variance $\g$. We will discuss Entropy-SGD~\citep{chaudhari2016entropy} in~\cref{ss:lesgd} which is an algorithm to solve
\beq{
    \argmin_x\ f_\g(x).
    \label{prob:le}
}
The $\parle$ algorithm instead solves
\beq{
    \argmin_{x,\ x^1,\ \ldots,\ x^n}\ \sum_{a=1}^n\ f_\g(x^a) + \f{1}{2 \r}\ \norm{x^a - x}^2.
    \label{prob:parle}
}
As we will see in~\cref{ss:equivalence}, for a semi-convex function $f(x)$, the three problems above, namely, Elastic-SGD~\cref{prob:esgd}, Entropy-SGD~\cref{prob:le} and $\parle$~\cref{prob:parle} are equivalent. They differ in their communication requirements: Entropy-SGD does not involve any communication, Elastic-SGD involves a large communication overhead while $\parle$ strikes a balance between the two.

$\parle$ can be thought of as running Entropy-SGD to minimize $f_\g(x^a)$ and coupling each member of the ensemble $x^a$ via Elastic-SGD. The loss function forces every $x^a$ to minimize its loss $f_\g(x^a)$ and also forces the reference $x$ to move towards the mean of all the $x^a$s. As training progresses, we let both $\g \to 0$ and $\r \to 0$. This technique is called ``scoping'' and it causes all replicas to collapse to a single configuration $x^*$ which is our solution. The use of this technique for Elastic-SGD is novel in the literature and it performs very well in our experiments (\cref{s:expts}).

\subsection{Motivation}
\label{ss:motivation}

\begin{wrapfigure}{r}{0.25\textwidth}
\centering
\includegraphics[width=0.25\textwidth]{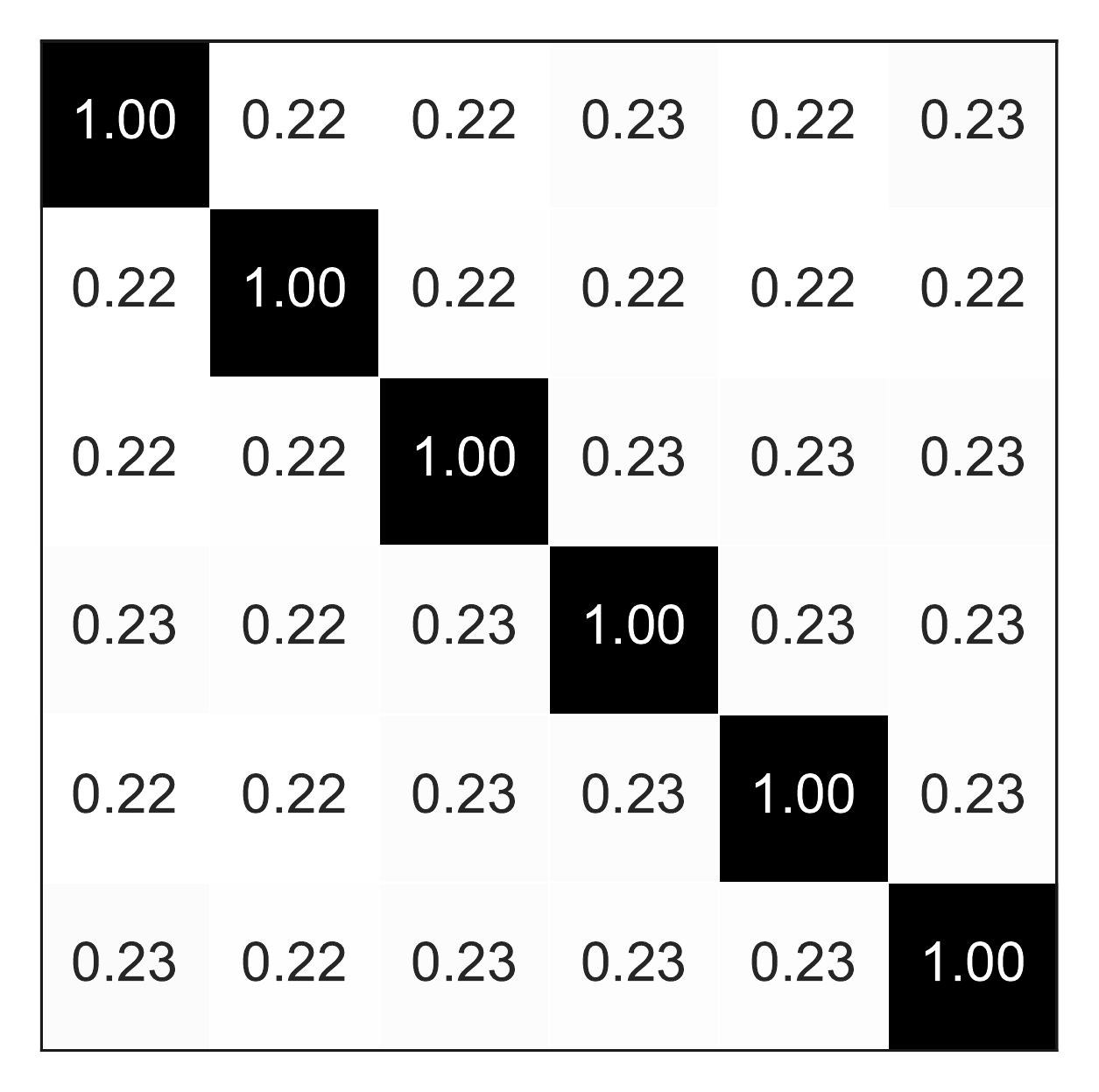}
\caption{\small Permutation invariant overlap of independently trained networks}
\label{fig:perdist}
\vspace*{-0.1in}
\end{wrapfigure}

Training independent copies of a deep network in parallel is difficult. Let us discuss an experiment that explores averaging such models and motivates the choice of the loss function in~\cref{ss:approach}. We trained $6$ instances of the $\allcnn$ architecture of~\citet{springenberg2014striving} on the CIFAR-10 dataset. These networks converge to a training error of $4.08 \pm 0.9\%$ and a validation error of $8.04 \pm 0.16\%$. Averaging their softmax predictions performs only slightly better than any individual network and the ensemble gets a validation error of $7.84\%$, indeed if we look at the correlation of the softmax predictions of these networks, it shows that they make mistakes on the same examples. This marginal improvement comes at the cost of a large test-time performance penalty.

On the other hand, a model that consists of the average weights of these independent networks (``one shot averaging'') performs poorly: it obtains $89.9\%$ validation error, which is equivalent to guessing. This is expected, given the non-convexity of the loss function for deep networks. Although individual networks might converge to good local minima, their average  need not even be locally optimal.

A typical deep network possess permutation symmetries, i.e., if the first and last layers are fixed, filters on intermediate layer can be permuted without changing the predictions. For a fully-connected network, this is akin to permuting the rows and columns of successive weight matrices. Post-training, we aligned each network to the first one by permuting its filters using a greedy layer-wise matching algorithm.~\cref{fig:perdist} shows the average overlap of the weights of these aligned filters; it can be seen as a permutation invariant metric between two networks. As we can see, these copies are very far away from each other in weight space.

What is surprising is that while a naively averaged model has close to $90\%$ validation error, a model obtained by averaging after aligning the weights performs much better at $18.7\%$ validation error. This suggests that if we can force the copies to remain aligned to each other during training, we can obtain one single average model at the end that combines these copies. $\parle$ uses a quadratic distance to align two copies during training. The loss function in~\cref{prob:parle} ensures that two replicas $x^a$ and $x^b$ are aligned through the quadratic forcing term $\sum_{a=1}^n\ \norm{x^a - x}^2$. This encourages members of the ensemble to be close to each other in  Euclidean space. For a small enough $\r > 0$, these replicas are constrained to have a large overlap while still minimizing their individual loss functions $f(x^a)$. As $\r \to 0$ towards the end of training, the overlap goes to unity.

\section{Background and related work}
\label{s:background}

The loss function of a deep network $f(x)$ is a non-convex function, and there have been numerous efforts to understand the geometry of the energy landscape. The relevance of this literature to our work is that it has been observed empirically that ``flat minima'' (local extrema where most of the eigenvalues of the Hessian are near zero) yield parameters that produce classifiers with good generalization~\citep{hochreiter1997flat,baldassi2016multilevel}. Entropy-SGD~\citep{chaudhari2016entropy} which we describe in the next section, was designed specifically to seek such minima. In the following section, we also discuss Elastic-SGD~\citep{zhang2015deep} since the updates of $\parle$ to minimize~\cref{prob:parle} combine these two algorithms.

\subsection{Entropy-SGD}
\label{ss:lesgd}

Consider the sub-problem~\cref{prob:le} that involves minimizing $f_\g(x)$. The authors in~\citet{chaudhari2016entropy} constructed an SGD-based algorithm to solve this which involves an inner-loop that executes Markov chain Monte Carlo (MCMC) updates. This algorithm can be written as
\begin{subequations}
    \aeq{
        y_{k+1} &= y_k - \eta' \sqbrac{\grad f(y_k) + \f{1}{\g}\ (y_k - x_k)}
        \label{eq:lesgd_y}\\
        z_{k+1} &= \a\ z_k + (1-\a)\ y_{k+1}
        \label{eq:lesgd_z}\\
        x_{k+1} &=
        \begin{cases}
            x_k - \eta \rbrac{x_k - z_{k+1}}  & \trm{if}\ k/L\ \trm{is\ an\ integer;}\\
            x_k & \trm{else}.
        \end{cases}
        \label{eq:lesgd_x}
    }
\label{eq:lesgd}
\end{subequations}
We initialize $y_k$ to $x_k$ every time $k/L$ is an integer. Let us parse these updates further, note that the $y_k$ variable performs gradient descent on the original loss function $f(\cdot)$ with a ``proximal'' term $\g^{-1}\ (y_k - x_k)$ that ensures that successive updates of $y_k$ stay close to $x_k$. The $z_k$ variables maintain a running exponential average of $y_k$ and the $x_k$ updates uses this average as the gradient. It can be proved that these updates are equivalent to performing gradient descent on $f_\g(x)$~\citep{chaudhari2016entropy}. The system~\cref{eq:lesgd} is thus simply
\[
    x_{k+1} = x_k - \eta\ \grad f_\g(x_k);
\]
with the gradient $\grad f_\g(x_k) = \g^{-1} (x_k - \ag{y_k})$ where $\ag{y_k}$ denotes the average of the $y_k$ iterates; we use exponential averaging in~\cref{eq:lesgd_z}. We have introduced a slight modification in~\cref{eq:lesgd_y} as compared to~\citet{chaudhari2016entropy}, namely, we do not add any MCMC noise. In practice, the gradient of a deep network $\grad f(\cdot)$ is computed on mini-batches and is a noisy estimate of the true gradient, so there is already some inherent stochasticity in~\cref{eq:lesgd_y}.

\subsection{Elastic-SGD}
\label{ss:esgd}

Let us now discuss how to minimize the objective in~\cref{prob:esgd}. Consider a setting where the objective is split between different workers $1 \leq a \leq n$, these could be physically different computers, or different process on the same machine. Each worker performs SGD to minimize its own objective while communicating the ``elastic'' gradient $\r^{-1} \rbrac{x_k^a - x_k}$ after each SGD update. The updates therefore look like
\begin{subequations}
    \aeq
    {
        x_{k+1}^a &= x_k^a - \eta\ \sqbrac{\grad f(x_k^a) + \f{1}{\r}\ \rbrac{x_k^a - x_k}} \quad \forall\ a \leq n
        \label{eq:esgd_xa}\\
        x_{k+1} &= x_k - \eta \rbrac{x_k - \f{1}{n}\ \sum_{a=1}^n\ x^a},
        \label{eq:esgd_x}
    }
\label{eq:esgd}
\end{subequations}
for all replicas $x^a$ with $a \leq n$. If one considers the variable $x$ as the ``master'' or the parameter server, we can see that these updates again have the flavor of the master updating itself with the average of the replicas. Indeed~\cref{eq:lesgd_y} and~\cref{eq:esgd_xa} are similar; while the former takes $L$ steps in the $y_k$ variable, the latter performs $n$ independent steps in the $x_k^a$ variables.

\subsection{Equivalence of Entropy-SGD and Elastic-SGD}
\label{ss:equivalence}

The resemblance of~\cref{eq:lesgd} and~\cref{eq:esgd} is not a coincidence, the authors in~\citet{chaudhari2017deep} proved that Elastic-SGD is equivalent to Entropy-SGD if the $y_k$ updates converge quickly, i.e., if the sub-objective of~\cref{eq:lesgd_y},
\[
    f(y) + \f{1}{2 \g} \norm{y - x}^2
\]
is strictly convex in $y$. This happens if $\grad^2 f(y) + \g^{-1}\ I \succ 0$. This condition implies that the stochastic process of the $y_k$ updates in~\cref{eq:lesgd_y} has an ergodic steady-state distribution whereby temporal averages ($z_k$ updates of~\cref{eq:lesgd_z}) are equivalent to spatial averages~\cref{eq:esgd_x}. Using different techniques, the two objectives can be shown equivalent under different approximations~\citep{baldassi2016unreasonable}.

Operationally speaking, Entropy-SGD is a sequential MCMC algorithm and hence hard to parallelize. The $y_k$ updates in~\cref{eq:lesgd_y} form a single trajectory of $L$ steps before the $x_k$ update~\cref{eq:lesgd_x} and it is difficult to execute chunks of this trajectory independently. On the other hand, Elastic-SGD is a naturally parallelizable algorithm but suffers from a large communication overhead; every weight update~\cref{eq:esgd_x} requires a reduce operation from all the workers and another broadcast of $x_{k+1}$ to each of them. This becomes prohibitive for large deep networks. Fortunately, the two algorithms are equivalent, and we can exploit this to minimize~\cref{prob:parle}.

\subsection{Scoping}
\label{ss:scoping}

We can see from~\cref{eq:le} that local entropy converges to the original loss function as $\g \to 0$ and it converges to a constant over the entire parameter space when $\g \to \infty$. This is also true for the loss function of Elastic-SGD
\[
    \sum_{a=1}^n\ f(x^a) + \f{1}{2 \r}\ \norm{x^a - x}^2;
\]
as $\r \to 0$, the variable $x$ converges to the average of the global minimizers of $f(x)$. For a convex loss $f(x)$, the replicas $x^1, \ldots, x^n$ and the reference $x$ all collapse to the same configuration. Our use of this technique for Elastic-SGD is novel in the literature and it performs very well in our experiments (\cref{s:expts}).

\subsection{Related work}
\label{ss:related work}

There are two primary ways to parallelize the training of deep networks. The first, called \textbf{model parallelism}, distributes a large model across multiple GPUs or compute nodes~\citep{krizhevsky2014one,dean2012large}. This does not scale well because intermediate results of the computation need to be transmitted to different compute nodes quickly, which puts severe limits on the latencies that the system can tolerate, e.g., large neural networks with recurrent or fully-connected layers are not amenable to model parallelism.

\textbf{Data parallelism} is more widely used~\citep{jin2016scale,moritz2015sparknet,taylor2016training} and maintains multiple copies of the network on different GPUs or compute nodes. Each mini-batch is split evenly amongst these copies who compute the gradient on their samples in parallel. A parameter server then aggregates these partial gradients and updates the weights. This aggregation can be either synchronous or asynchronous. The former is guaranteed to be exactly equivalent to a non-distributed, sequential implementation while the latter can work with large communication latencies but offers guarantees only for convex or sparse problems~\citep{recht2011hogwild,duchi2013estimation}.

Both \textbf{synchronous and asynchronous gradient aggregation} necessitate a high per-worker load~\citep{qi2016paleo} if they are to scale well. This is difficult to do in practice because small mini-batches require frequent communication while large batch-sizes suffer from a degradation of the generalization performance. Previous work in the literature has mostly focused on minimizing the communication requirements using stale gradients in HogWild! fashion~\citep{recht2011hogwild,zhang2015staleness}, extreme quantization of gradients~\citep{seide20141} etc. These heuristics have shown impressive performance although it is hard to analyze their effect on the underlying optimization for deep networks. Synchronous approaches often require specialized hardware and software implementations to hide communication bottlenecks~\citep{wu2015deep,tensorflow2015-whitepaper,chen2015mxnet}. Recent work by~\citet{goyal2017accurate} shows that one can indeed obtain generalization performance comparable to small batch-sizes by a careful tuning of the learning rate. While large batches, in theory, can enable large learning rates, as is often seen in practice, optimization of a deep network with a large learning rate is difficult and even the training loss may not decrease quickly. The authors demonstrate that a warm-up scheme that increases the learning rate from a small initial value followed by the usual annealing leads to similar training and validation error curves as those of small mini-batches. \textbf{$\parle$ can benefit from both synchronous and asynchronous gradient aggregation}, indeed, each replica in~\cref{prob:parle} can itself be data-parallel. Our work shows that exploiting the properties of the optimization landscape leads to better generalization performance.

Ensemble methods train multiple models and average their predictions to improve generalization performance\footnote{~the top-performing methods on ImageNet are ensembles of deep networks (\href{http://image-net.org/challenges/LSVRC/2016/results}{http://image-net.org/challenges/LSVRC/2016/results})}. Deploying ensembles of deep networks in practice however remains challenging due to both the memory footprint and test-time latency of state-of-the-art networks. For instance, an ensemble of an object detection network like YOLO~\citep{redmon2016you} that works at $45$ frames per second on PASCAL VOC can no longer be used in a real-time environment like a self-driving car~\citep{leonard2008perception} with limited computational budget. In contrast to an ensemble, \textbf{$\parle$ results in one single model that performs well at test-time}. Second, training multiple models of an ensemble is computationally expensive~\citep{loshchilov2016sgdr,huang2017snapshot}, and as our experiment in~\cref{ss:motivation} shows, the ensemble obtains a marginal improvement over each individual model. \textbf{$\parle$ maintains a correlated, robust ensemble} during training and returns the average model that obtains better errors than a naive ensemble.

\section{Parle}
\label{s:parle}

Stochastic gradient descent step to minimize~\cref{prob:parle} amounts to combining Entropy-SGD in~\cref{eq:lesgd} and Elastic-SGD in~\cref{eq:esgd}. For all replicas $a \leq n$, $\parle$ performs the following updates:
\begin{subequations}
    \aeq{
        y_{k+1}^a &= y_k^a - \eta' \sqbrac{ \grad f(y^a_k) + \f{1}{\g} \rbrac{y_k^a - x_k^a}}{}
        \label{eq:parle_y}\\
        z_{k+1}^a &= \a\ z_{t}^a + (1-\a)\ y_{k+1}^a
        \label{eq:parle_z}\\
        x_{k+1}^a &=
        \begin{cases}
            x_k^a - \eta \rbrac{x_k^a - z_{k+1}^a}  - \f{\eta}{\r}\ \rbrac{x_k^a - x_k} & \trm{if}\ k/L\ \trm{is\ an\ integer}\\
            x_k^a & \trm{else},
        \end{cases}
        \label{eq:parle_xa}\\
        x_{k+1} &=
        \begin{cases}
            x_k - \f{\eta''\ n}{\r} \rbrac{ x_k - \f{1}{n}\ \sum_{a=1}^n x_k^a} & \trm{if}\ k/L\ \trm{is\ an\ integer}\\
            x_k & \trm{else}.
        \end{cases}
        \label{eq:parle_x}
    }
\label{eq:parle}
\end{subequations}
We reset $y_k^a$ to $x_k^a$ every time $k/L$ is an integer. Notice that~\cref{eq:parle_y} and~\cref{eq:parle_z} are the same as~\crefrange{eq:lesgd_y}{eq:lesgd_z}. The update for $x_k$ in~\cref{eq:parle_x} is also the same as the update for the reference variable $x_k$ in~\cref{eq:esgd} with a step-size of $\r$. The only difference is that~\cref{eq:parle_xa} takes a gradient step using both the gradient of $f_\g(x^a) \propto (x_k^a - z_k^a)$ and the gradient of the elastic term $\r^{-1} \rbrac{x_k^a - x_k}$.

\begin{remark}[\textbf{Scoping and learning rate annealing}]
The gradient of local entropy is
\[
    \grad f_\g(x^a_k) = \g^{-1}(x^a_k - z^a_{k+1}).
\]
The learning rate $\eta$ for a deep network is typically reduced in steps as the training progresses. As discussed in~\cref{ss:lesgd}, we would like to take $\g \to 0$, which interferes with this learning rate annealing. We have therefore scaled up the learning rate $\eta$ by $\g$ in~\cref{eq:parle_xa}. We do not scale it for the second term $\rbrac{x_k^a - x_k}$ because this automatically gives us a weighted combination of the gradients of local entropy and the proximal term. This choice is akin to picking a modified annealing schedule for the parameters $\g$ and $\r$.
\end{remark}

\subsection{Hyper-parameter choice}
\label{ss:hyper}
For all the experiments in this paper, the parameters of $\parle$ in~\cref{eq:parle} are fixed to $L=25$, $\a = 0.75$. We also set $\eta'' = \r/n$ in~\cref{eq:parle_x}, i.e., at each update we simply average the replicas to get
\[
    x_{k+1} = \f{1}{n}\ \sum_{a=1}^n x_k^a.
\]
The parameters $\g$ and $\r$ are updated every time $k/L$ is an integer to
\beq{
    \g_k = \g_0 \rbrac{1- \f{1}{2 B}}^{\floor{k/L}} \quad \trm{and}\ \quad
    \r_k = \r_0 \rbrac{1- \f{1}{2 B}}^{\floor{k/L}},
    \label{eq:scoping}
}
where $\g_0 = 10^2$ and $\r_0 = 1$ and $B$ is the number of mini-batches in the dataset. We clip $\g$ at $1$ and $\r$ at $0.1$. The parameter $\eta'$ is fixed to be the initial learning rate.

The only remaining parameter is the learning rate $\eta$ which we drop by a factor of $5-10$ when the validation error plateaus,~\cref{s:expts} provides more details. We have found that this algorithm is quite robust to parameter changes. In particular, both the speed of convergence and the final generalization error are insensitive to the exact values of $\g_0$ or $\r_0$.

\begin{remark}[\textbf{Nesterov's momentum}]
We use Nesterov's momentum (fixed to $0.9$) for updating the variables $y^a_k$ in~\cref{eq:parle_y} and $x_k^a$ in~\cref{eq:parle_xa}. With our specific choice of $\eta'' = \r/n$, the $x_k$ update does not have first-order dynamics, and we therefore do not use momentum to update it. With other choices of $\eta''$, one could use momentum for~\cref{eq:parle_x} as well, but we found this to converge marginally slower.
\end{remark}

\subsection{Many deputies under one sheriff}

Consider an alternative optimization problem:
\beq{
    \argmin_{x,\ x^1, \ldots,\ x^n,\ y^1, \ldots,\ y^n}\ \sum_{a=1}^n\ \rbrac{\sum_{b=1}^n\ f(y^b) + \f{1}{2 \g}\ \norm{y^b - x^a}^2} + \f{1}{2 \r}\ \norm{x^a - x}^2.
    \label{prob:sheriff}
}
Under an ergodicity assumption, from the discussion in~\cref{s:background}, this is equivalent to~\cref{prob:parle}. It is also equivalent to~\cref{eq:esgd} with a modified coupling between the workers where the constant $\r$ in~\cref{prob:esgd} takes to different values. This loss function provides a completely distributed version of $\parle$ and has the interpretation of ``deputies'' $x^a$ connected to a ``sheriff'' $x$ through the term $\f{1}{2 \r} \norm{x^a - x}^2$. Each deputy in turn controls the workers $y^b$ through the proximal term $\f{1}{2 \g} \norm{y^a - x^a}^2$.

The variables $x, x^a, y^b \in \reals^N$ are high-dimensional and there are $n^2$ copies. Optimizing~\cref{prob:sheriff} therefore involves a large communication overheard: each deputy communicates $\OO(2 n N)$ bits with its workers and the sheriff communicates another $\OO(2 n N)$ bits with each deputy. The total communication complexity is thus quadratic $\OO(n^2 N^2)$ at each weight update which is prohibitive for large $N$. On the other hand, using the updates in~\cref{eq:lesgd} and~\cref{eq:esgd} to minimize~\cref{prob:parle} results in an amortized communication complexity of $\OO \rbrac{2 n N/L}$ while remaining equivalent to~\cref{prob:sheriff}.

\begin{remark}[\textbf{Running Parle on diverse computational platforms}]
We can extend the above loss function further and again replace $f(y^b)$ with either $f_\d(y^b)$ or even the Elastic-SGD loss function. Coupled with the idea of HogWild! or stale gradients, one can achieve a seamless trade-off between replicas that may have a lot of computational budget but relatively scarce communication resources, e.g., GPUs, and replicas that can communicate faster than they can compute, e.g., CPUs and mobile devices. Entropy-SGD is naturally suited to the former and Elastic-SGD is naturally suited to the latter, the loss function of $\parle$ shows how to couple such diverse computational platforms together.
\end{remark}

\section{Empirical validation}
\label{s:expts}

This section discusses experimental results on a variety of benchmark image classification datasets, namely, MNIST (\cref{ss:mnist}), CIFAR-10 and CIFAR-100 (\cref{ss:cifar}) and SVHN (\cref{ss:svhn}). $\parle$ obtains nearly state-of-the-art errors with a significant wall-clock time speed up without any additional hyper-parameters. In~\cref{s:split_data}, we show that $\parle$ obtains error rates better than SGD with full data even in the case when each replica only has access to a subset of the data. For all the following experiments, we use SGD with Nesterov's momentum (fixed to $0.9$) as a baseline and compare the performance of $\parle$, Entropy-SGD and Elastic-SGD. We compute the mean and standard deviation of the validation error over $3$ runs for each algorithm with random initialization. However, for some networks that take a long time to train, we only report the results of a single run (see~\cref{tab:expts} and~\cref{tab:split_data}).

\subsection{Implementation details and communication latencies}
\label{ss:communication}

\begin{wrapfigure}{r}{0.45\textwidth}
\centering
\includegraphics[width=0.45\textwidth]{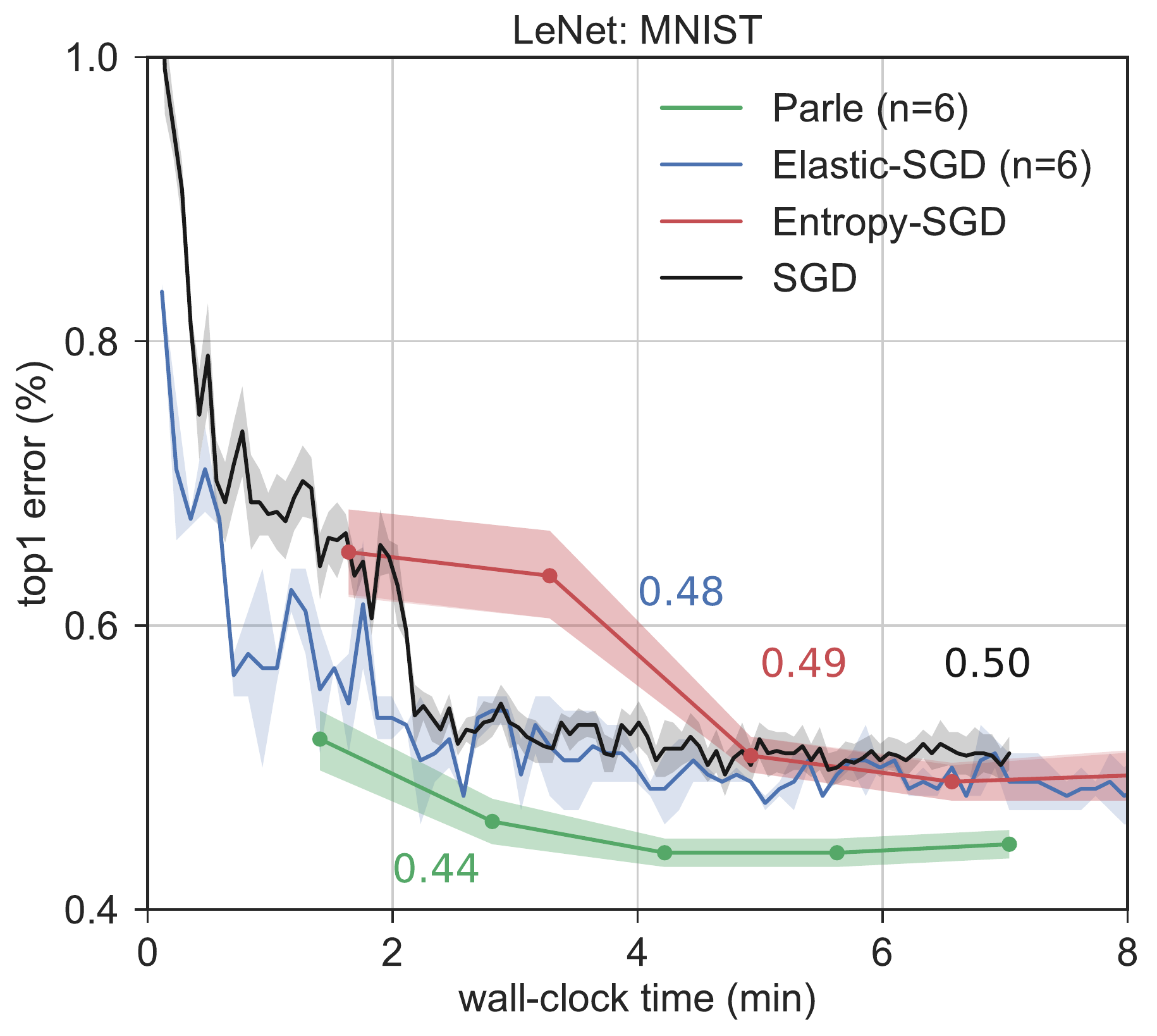}
\caption{\small Validation error: $\lenet$ on MNIST}
\label{fig:lenet_full_valid}
\end{wrapfigure}

We have implemented a parallel version of the updates in~\cref{eq:parle}, i.e., we execute multiple neural networks in the same process on a standard desktop with $3$ GPUs. The communication with the master for the $x_k$ variable happens via optimized NCCL\footnote{\href{https://devblogs.nvidia.com/parallelforall/fast-multi-gpu-collectives-nccl}{https://devblogs.nvidia.com/parallelforall/fast-multi-gpu-collectives-nccl}} routines on PCI-E. This is implemented using PyTorch\footnote{\href{http://pytorch.org}{http://pytorch.org}} in Python but it scales well to $n \approx 16$ replicas. In particular, the reduce operation required in~\crefrange{eq:parle_xa}{eq:parle_x} does not incur a large communication overhead. For instance, on our machine, each mini-batch of size $128$ for the wide residual network $\wrncifar$ in~\cref{ss:cifar} takes $528$ ms on an average, while steps~\crefrange{eq:parle_xa}{eq:parle_x} take $2.8$ ms, their ratio is $0.52\%$ and is therefore negligible. For the $\allcnn$ network in~\cref{s:split_data}, this ratio is $0.43\%$. Let us however note that addressing communication bottlenecks in a distributed implementation is non-trivial. As an example, the authors in~\citet{goyal2017accurate} perform gradient aggregation and backprop in parallel. $\parle$ can also benefit from such techniques although that is not the focus of our paper.

\begin{remark}[\textbf{Plotting against wall-clock time}]
Entropy-SGD and SGD are sequential algorithms while $\parle$ and Elastic-SGD can run on all available GPUs simultaneously. In order to obtain a fair comparison, we run the former two in data-parallel fashion on three GPUs and plot all error curves against the wall-clock time. For the networks considered in our experiments, with a batch-size of $128$, the efficiency of a data-parallel implementation in PyTorch is above $90\%$ (except for $\lenet$).
\end{remark}

\subsection{MNIST}
\label{ss:mnist}

We use the standard $\lenet$ architecture~\citep{lecun1998gradient} with ReLU nonlinearities~\citep{nair2010rectified}, batch-normalization~\citep{ioffe2015batch}. It has two convolutional layers with $20$ and $50$ channels, respectively, followed by a fully-connected layer of $500$ hidden units that culminates into a $10$-way softmax. Both the convolutional and fully-connected layers use a dropout~\citep{srivastava2014dropout} of probability $0.25$. We do not perform any pre-processing for MNIST. The learning rate is initialized to $0.1$ and dropped by a factor of $10$ at epochs $[30,60,90]$ for SGD and only once, after the second epoch, for Entropy-SGD and $\parle$.
\cref{fig:lenet_full_valid} shows the validation errors for $\lenet$, $\parle$ obtains a validation error of $0.44 \pm 0.01\%$ with three replicas as compared to about $0.48$-$0.5\%$ on $\lenet$ with SGD, Entropy-SGD and Elastic-SGD.

\subsection{CIFAR-10 and CIFAR-100}
\label{ss:cifar}

We use the $\wrncifar$ network of~\citet{zagoruyko2016wide} for these datasets; this architecture was shown to have very good empirical performance in spite of not being very deep. We use the same training pipeline and hyper-parameters as that of the original authors to enable an easy comparison. In particular, we perform global contrast normalization followed by ZCA normalization and train with data-augmentation which involves random mirror-flipping (with probability $0.5$) and random crops of size $32 \times 32$ after padding the image by $4$ pixels on each side. We use a dropout of probability $0.3$ and weight decay of $5\times 10^{-4}$. The learning rate is initialized to $0.1$ and dropped by a factor of $5$ at epochs $[60,120,180]$ for SGD and $[2,4,6]$ for Entropy-SGD and $\parle$. The former is taken from the original paper while the later was constructed using the heuristic that $\parle$ and Entropy-SGD use $L=25$ gradient evaluations per weight update.

\begin{figure}[!htpb]
\centering
    \begin{subfigure}[t]{0.49\textwidth}
        \includegraphics[width=\textwidth]{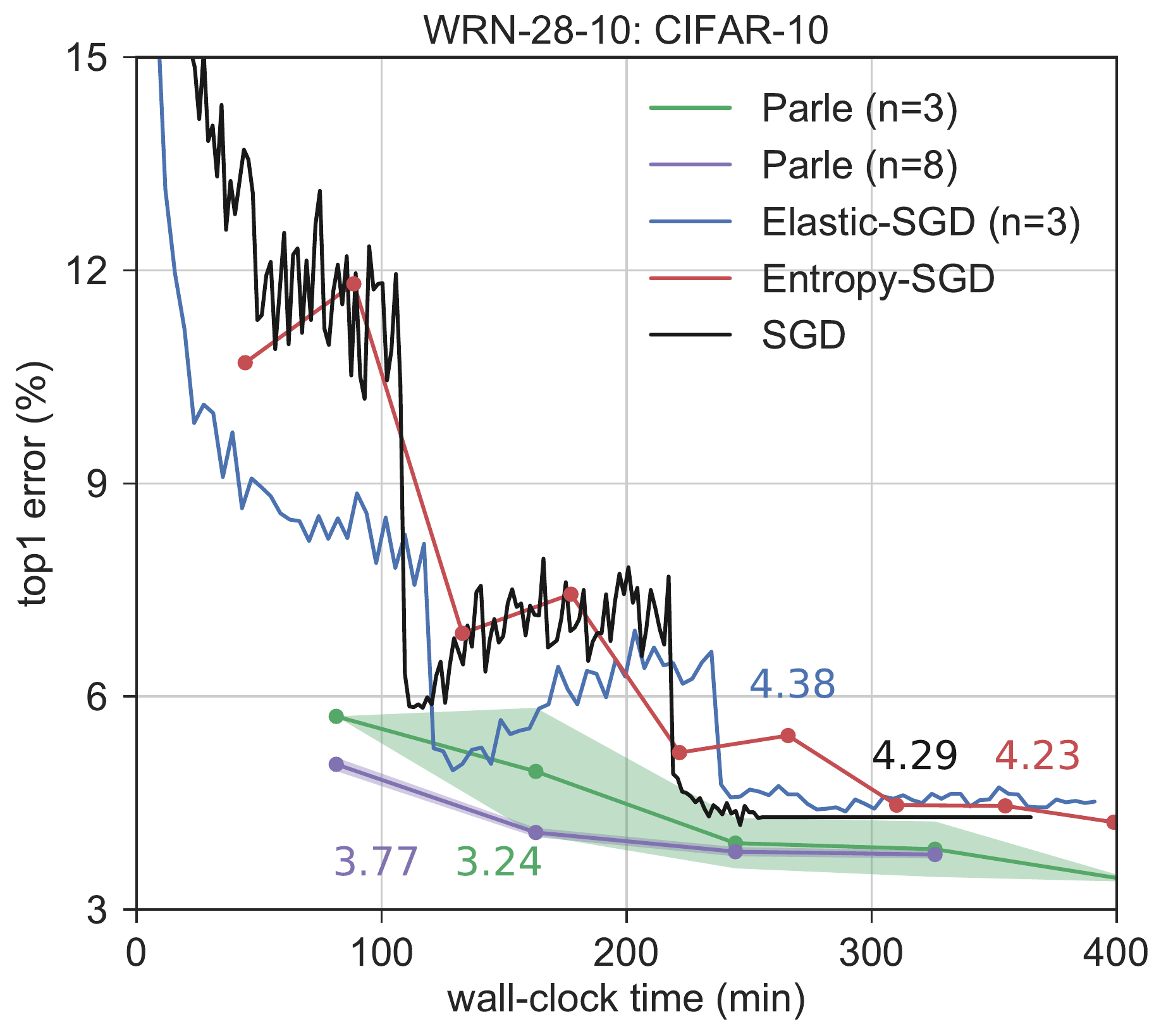}
        \caption{}
        \label{fig:wrn_cifar10_full_valid}
    \end{subfigure}
    \begin{subfigure}[t]{0.49\textwidth}
        \includegraphics[width=\textwidth]{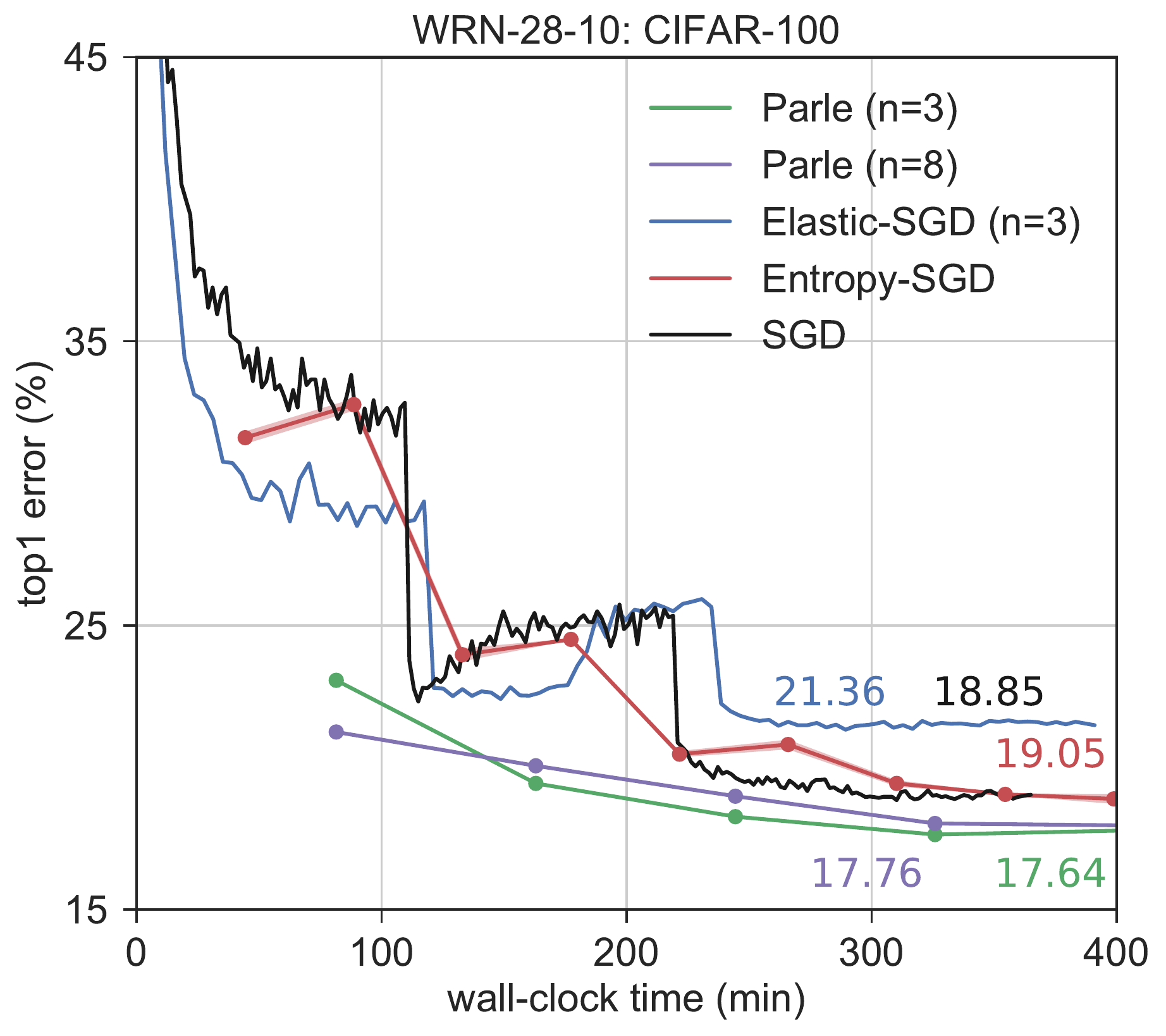}
        \caption{}
        \label{fig:wrn_cifar100_full_valid}
    \end{subfigure}
\label{fig:wrn_cifar}
\caption{\small Validation error of $\wrncifar$ on CIFAR-10 (\cref{fig:wrn_cifar10_full_valid}) and CIFAR-100 (\cref{fig:wrn_cifar100_full_valid})}
\end{figure}

\crefrange{fig:wrn_cifar10_full_valid}{fig:wrn_cifar100_full_valid} show the empirical performance on CIFAR-10 and CIFAR-100 respectively with the same data summarized in~\cref{tab:expts}. In our implementation, we obtained a validation error of $4.29\%$ with SGD as compared to $3.89\%$ by~\citet{zagoruyko2016wide} for CIFAR-10, however our baseline for CIFAR-100 matches well with theirs, both obtain $18.85\%$ validation error. On CIFAR-10, $\parle$ with $n=3$ replicas obtains a significantly better validation error of $3.24\%$ while it obtains $17.64\%$ error on CIFAR-100. Note that these are both more than $1\%$ better than the baseline SGD with exactly the same network and pre-processing. For these datasets, we found that the benefit of adding more replicas is small, with $n=8$, we see an initial speed up for both CIFAR-10 and CIFAR-100, but the network converges to a worse error with the same hyper-parameters. Note that if more GPUs are available, each replica can itself be run in a data-parallel fashion to further accelerate training time. Nevertheless, both versions of $\parle$ are better than the baseline SGD implementation.

It is instructive to compare the performance of Elastic-SGD and Entropy-SGD with $\parle$. Since $\parle$ essentially combines these two algorithms, as we saw in~\cref{s:parle}, it is a more powerful than either of them. This is corroborated by our experimental evidence. We also observed that Entropy-SGD obtains very similar errors as those of Elastic-SGD with scoping on the $\r$ parameter. Scoping was adapted from the results of~\citet{chaudhari2016entropy} and in our experience, it improves the performance of Elastic-SGD significantly.

Our errors on CIFAR-10 and CIFAR-100 are better than those reported previously in published literature for a single model~\footnote{~Recent work by~\citet{gastaldi2017shake} reports an error of $2.8\%$ on CIFAR-10 and $15.85\%$ on CIFAR-100 using ``shake-shake'' regularization on a three-branch residual network that is trained for $1800$ epochs}. In fact, our result on CIFAR-10 is better than the $3.44\%$ reported error in~\citet{huang2017snapshot} on an ensemble of six DenseNet-100 networks~\citep{huang2016densely}. Our result on CIFAR-100 is only slightly worse than a DenseNet-100 ensemble which gets $17.41\%$~\citep{huang2017snapshot}.

{
\setlength{\heavyrulewidth}{1.5pt}
\renewcommand{\arraystretch}{1.65}
\begin{table}[H]
\centering
\resizebox{\columnwidth}{!}
{
\begin{tabular}{p{5cm} | c c | c c | c c | c c}
\toprule
\rowcolor{gray!15} Model & \multicolumn{2}{c}{$\parle$} & \multicolumn{2}{c}{Elastic-SGD} & \multicolumn{2}{c}{Entropy-SGD} & \multicolumn{2}{c}{SGD}\\
 \toprule
    & Error & Time & Error & Time & Error & Time & Error & Time\\[0.05in]
    \rowcolor{gray!15}
    $\lenet$ (MNIST, $n=6$)
    & $\bf{0.44 \pm 0.01}$ & $\bf{4.24}$
    & $0.48 \pm 0.01$ & $5$
    & $0.49 \pm 0.01$ & $6.5$
    & $0.5 \pm 0.01$ & $5.6$\\

    $\wrncifar$ (CIFAR-10, $n=3$)
    & $\bf{3.24 \pm 0.1}$ & $\bf{400}$
    & $4.38$ & $289$
    & $4.23$ & $400$
    & $4.29$ & $355$\\


    \rowcolor{gray!15}
    $\wrncifar$ (CIFAR-100, $n=3$)
    & $\bf{17.64}$ & $\bf{325}$
    & $21.36$ & $317$
    & $19.05 \pm 0.03$ & $400$
    & $18.85$ & $355$\\


    $\wrnsvhn$ (SVHN)
    & $1.68 \pm 0.01$ & $592$
    & $\bf{1.57}$ & $\bf{429}$
    & $1.64$ & $481$
    & $1.62$ & $457$\\

\bottomrule
\end{tabular}
}
\vspace*{0.1in}
\caption{\small Summary of experimental results: Validation error (\%) at wall-clock time (min)}
\label{tab:expts}
\end{table}

\subsection{SVHN}
\label{ss:svhn}

\begin{wrapfigure}{r}{0.49\textwidth}
\centering
\includegraphics[width=0.49\textwidth]{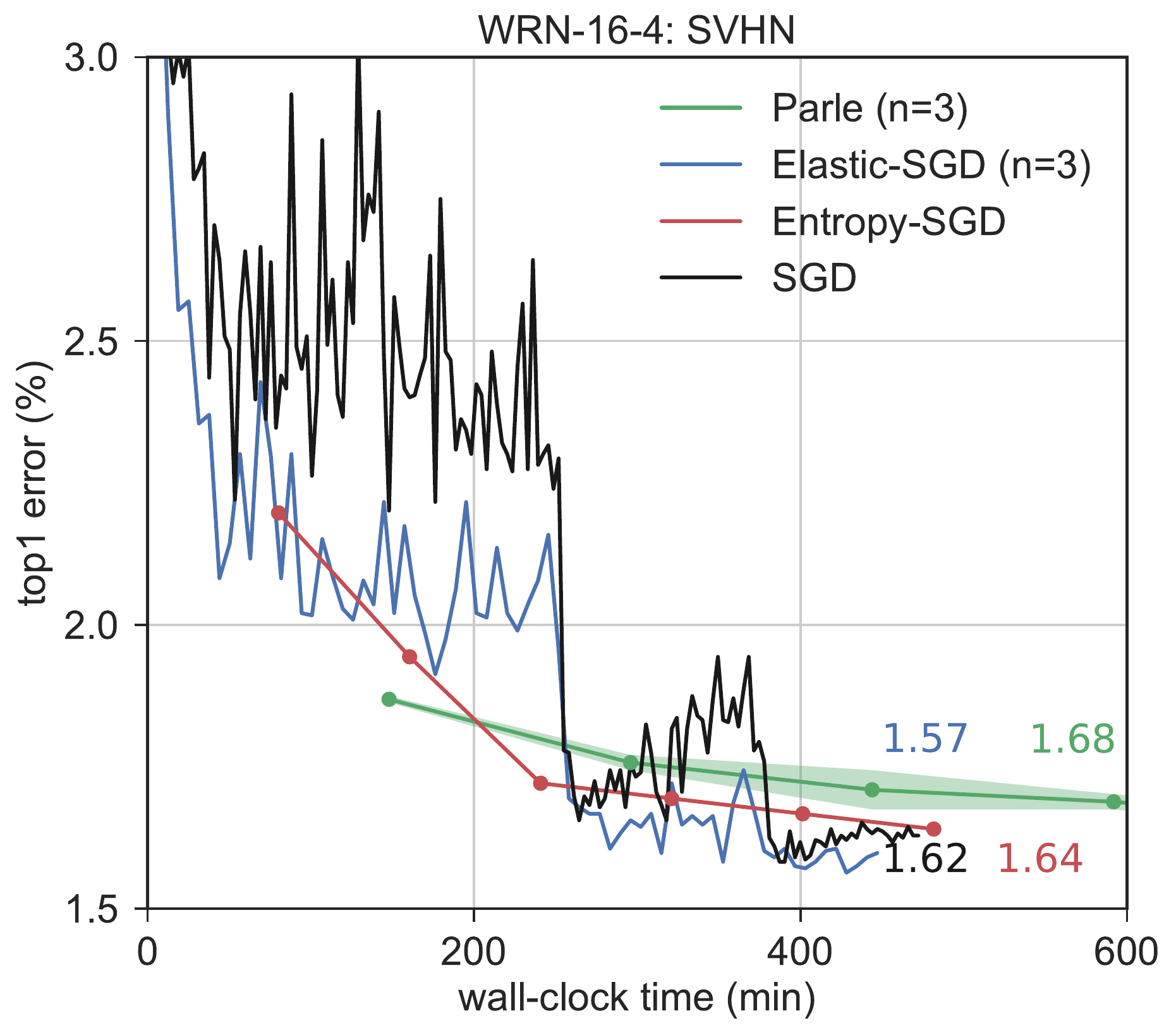}
\caption{\small Validation error: $\wrnsvhn$ on the SVHN dataset}
\label{fig:svhn_full_valid}
\end{wrapfigure}

SVHN is a dataset consisting of house numbers from Google's Street View and contains about $600,000$ images of digits. We use the $\wrnsvhn$ network of~\citep{zagoruyko2016wide}. We perform global contrast normalization for the input images and do not perform data augmentation. The dropout probability is set to $0.4$ and weight decay is set to $5 \times 10^{-4}$. The learning rate is initialized to $0.01$ and dropped by a factor of $10$ at epochs $[80,120]$ for SGD and epochs $[2,4]$ for Entropy-SGD and $\parle$.

\cref{fig:svhn_full_valid} shows the validation error for SVHN using $\parle$ with three replicas, Entropy-SGD, and SGD. For comparison, the authors in~\citet{zagoruyko2016wide} report $1.64\%$ error with SGD. All the three algorithms obtain comparable errors in this case, with Elastic-SGD being marginally better. For comparison, the best reported result on SVHN is using a larger network $\wrnsvhnl$ which gets $1.54\%$ error, this is very close to our result with Elastic-SGD with scoping. Let us note that Elastic-SGD does not work this well without scoping, we did not get errors below $1.9\%$ on SVHN.

\subsection{Training error}
\label{ss:training_error}

Let us look at the training error for $\wrncifar$ on CIFAR-10 (\cref{fig:wrn_cifar10_full_train}) and CIFAR-100 (\cref{fig:wrn_cifar100_full_train}) and $\wrnsvhn$ on SVHN (\cref{fig:wrn_svhn_full_train}). Note that while SGD and Elastic-SGD always converge to near-zero training errors, both Entropy-SGD and $\parle$ have much larger training error and do not over-fit as much. The minima discovered by SGD and $\parle$ are qualitatively different: while the former manages to get almost zero training error and converge to the global minimum of the loss function, it ends up over-fitting to the training data and does not generalize as well. $\parle$ obtains superior generalization performance at the cost of under-fitting to the training data. This also sheds light on the structure of the loss function. For the purposes of deep learning, with current regularization techniques, it is not always important to converge to the global optimum. Flat minima, which may exist at higher energy levels instead afford better generalization performance and can be found by algorithms like Entropy-SGD and $\parle$ that are designed to seek them. Note that Elastic-SGD only guaranteed to find flat minima for semi-convex loss functions~\citep{chaudhari2017deep}.

\begin{figure}[!htpb]
\centering
    \begin{subfigure}[t]{0.32\textwidth}
        \includegraphics[width=\textwidth]{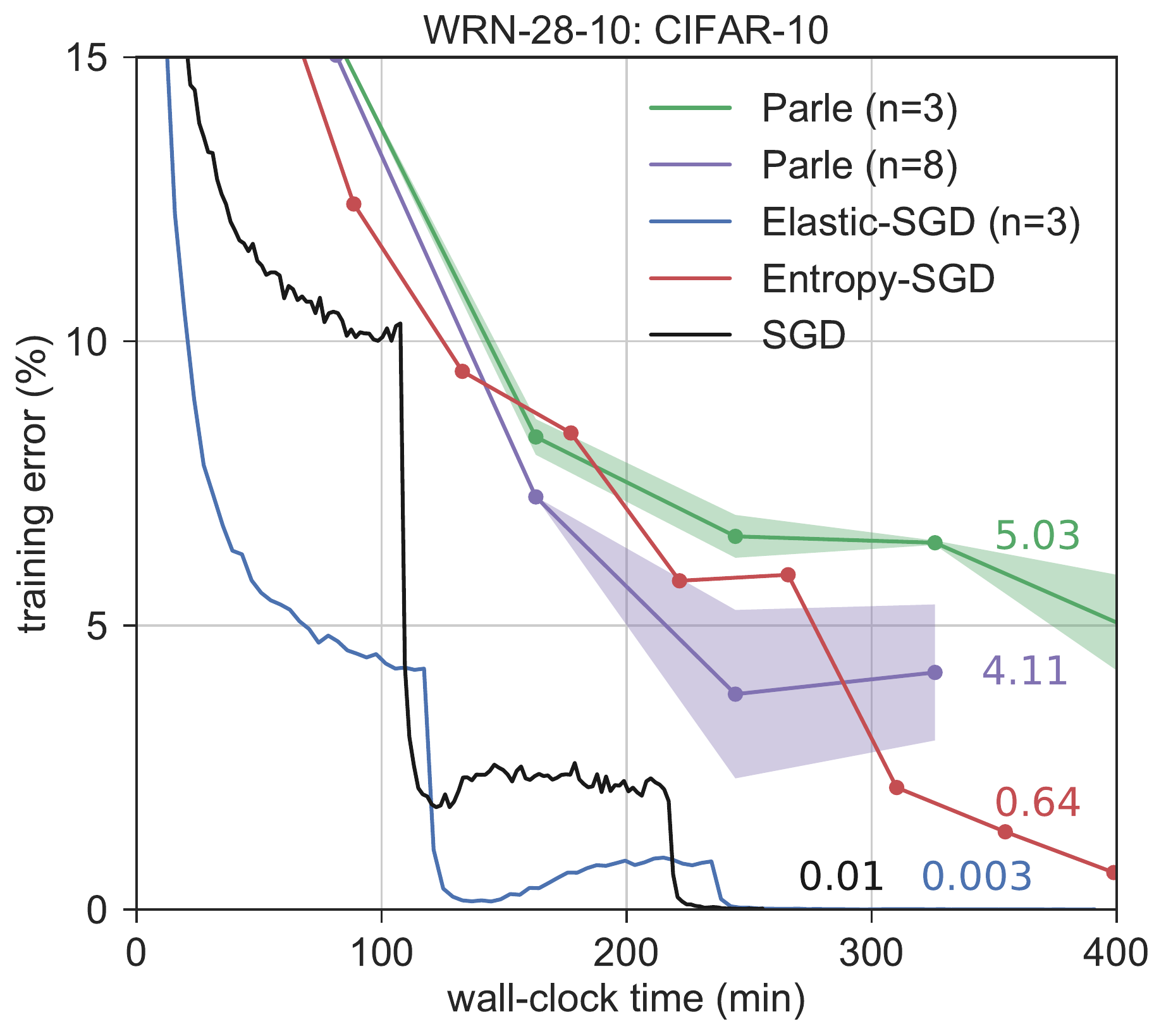}
        \caption{}
        \label{fig:wrn_cifar10_full_train}
    \end{subfigure}
    \begin{subfigure}[t]{0.32\textwidth}
        \includegraphics[width=\textwidth]{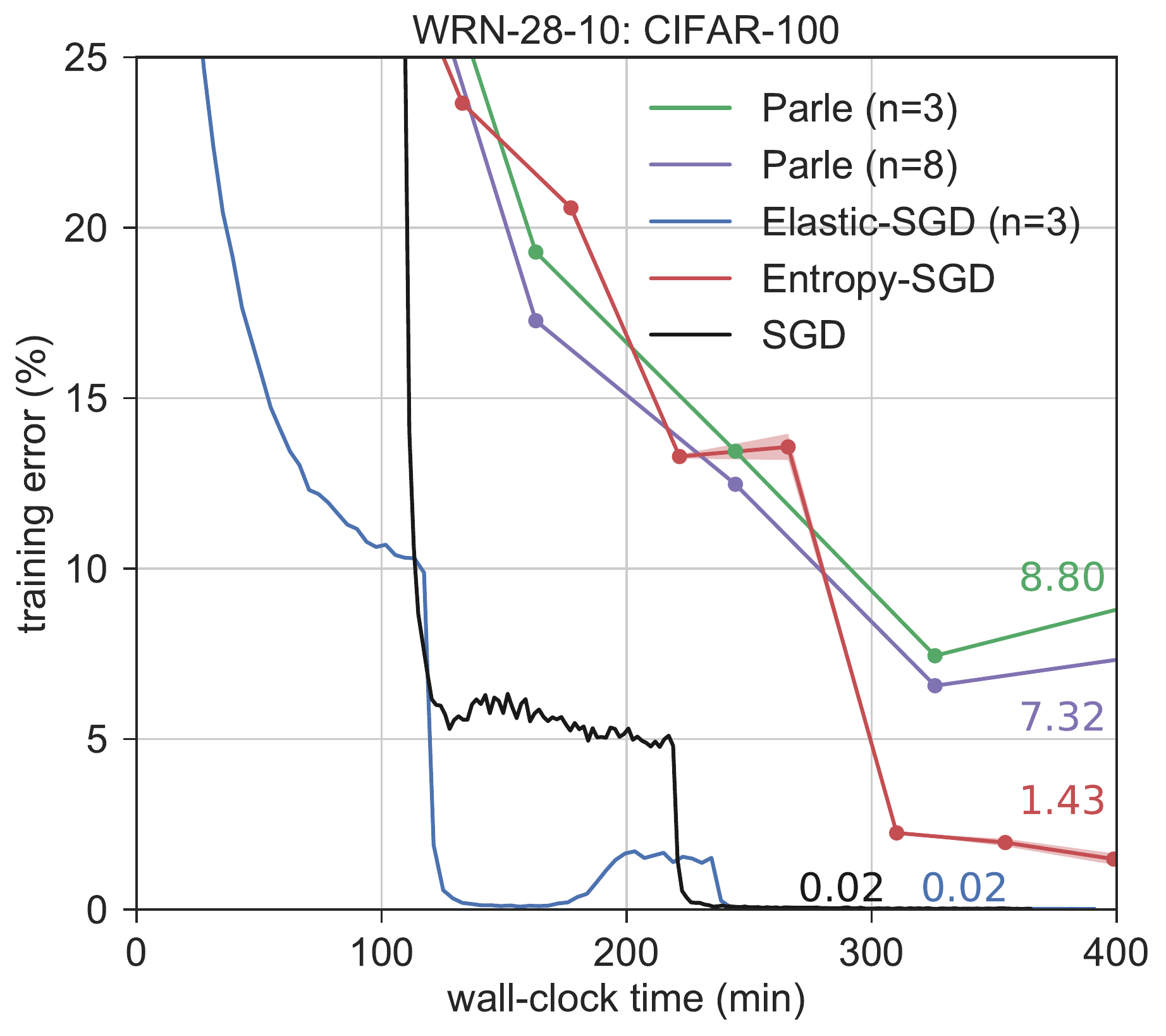}
        \caption{}
        \label{fig:wrn_cifar100_full_train}
    \end{subfigure}
    \begin{subfigure}[t]{0.32\textwidth}
        \includegraphics[width=\textwidth]{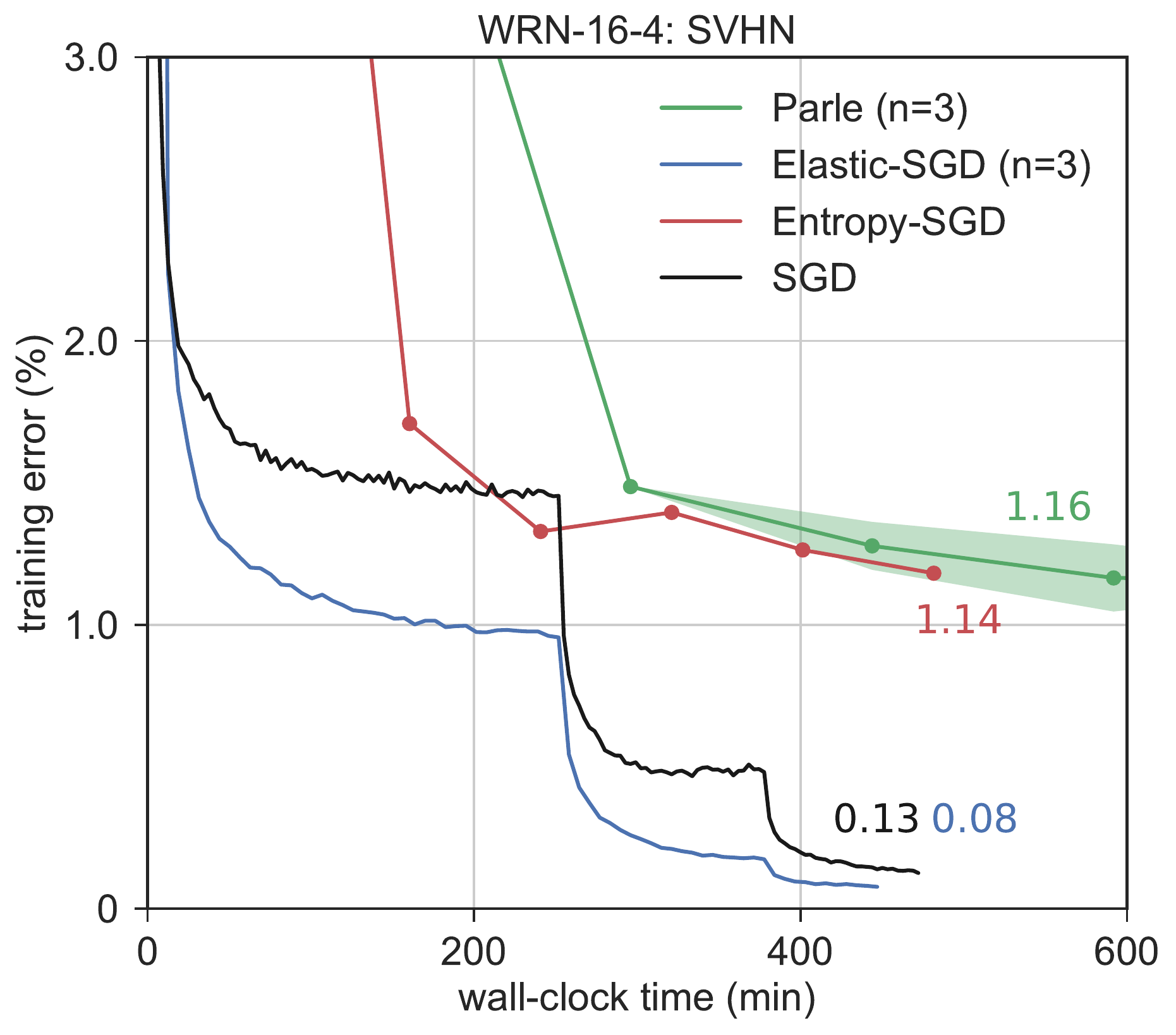}
        \caption{}
        \label{fig:wrn_svhn_full_train}
    \end{subfigure}
\caption{\small Training error on CIFAR-10 (\cref{fig:wrn_cifar10_full_valid}), CIFAR-100 (\cref{fig:wrn_cifar100_full_valid}) and SVHN (\cref{fig:wrn_svhn_full_train})}
\label{fig:wrn_train}
\end{figure}



\section{Splitting the data between replicas}
\label{s:split_data}

In the experiments above, each replica in $\parle$ has access to the entire dataset. Our aim in this section is to explore how much of this improvement can be traded off for speed. If we denote the entire dataset by $\xi$, each replica $x^a$ (see~\cref{s:parle}) only operates on a subset $\xi^a$. We split the dataset evenly amongst the replicas, i.e.,
\[
    \xi = \bigcup_{1 \leq a \leq n}\ \xi^a
\]
and all $\xi^a$ are of the same size. In particular, we ensure that each sample lies in at least one of the subsets $\xi^a$. By doing so, we would like to explore the efficacy of the proximal term $\f{1}{2 \r}\ \norm{x^a - x}^2$ in~\cref{prob:parle}. Effectively, the only way a replica $x^a$ gets a gradient term on $\xi^b$ is through this term. We will consider two cases, (i) with $n=3$ replicas and each gets $50\%$ of the training data, and (ii) with $n=6$ replicas and each gets $25\%$ of the training data.

We use the $\allcnn$ network of~\citet{springenberg2014striving} for these experiments on CIFAR-10. Again, the hyper-parameters are kept the same as the original authors, in particular, we use a dropout probability of $0.5$ with weight decay of $10^{-3}$ and data augmentation with mirror flips and random crops. Entropy-SGD and SGD are non-distributed algorithms and cannot handle this scenario, we therefore compare $\parle$ with Elastic-SGD (both operate on the same subsets) and a data-parallel SGD on three GPUs with access to the entire dataset. The latter is our baseline and obtains $6.15\%$ validation error. As~\cref{fig:allcnn_cifar10_half_valid} and~\cref{fig:allcnn_cifar10_fourth_valid} show, quite surprisingly, $\parle$ obtains better error than SGD in spite of the dataset being split between the replicas. The speedup in wall-clock time in~\cref{fig:allcnn_cifar10_fourth_valid} is a consequence of the fact that $\parle$ has very few mini-batches. Elastic-SGD with split data converges quickly but does not obtain a validation error as good as SGD on the full dataset. For comparison, $\parle$ obtains an error of $5.18\%$ on CIFAR-10 with this network if it has access to the entire dataset in $75$ minutes. To our knowledge, this is the best reported error on CIFAR-10 without a residual network, which is itself of importance since $\allcnn$ is about $9\times$ faster at test-time than $\wrncifar$. In~\cref{tab:split_data}, for comparison, we also report the error of SGD with access to a random subset of the training data (averaged over $3$ independent runs); as expected, this is much worse than its error with access to the full dataset.

This experiment shows that the proximal term $\f{1}{2 \r}\ \norm{x^a - x}^2$ is strong enough to pull the replicas towards good regions in the parameter space in spite of their individual gradients being computed on different datasets $\xi^a$. If the proximal term is strong enough (as $\r \to 0$), they can together converge to a region in the parameter space that works for the entire dataset. Exploiting this observation in a data distributed setting~\citep{mcmahan2016communication} to obtain state-of-the-art performance is a promising direction for future work.

{
\setlength{\heavyrulewidth}{1.5pt}
\renewcommand{\arraystretch}{1.5}
\begin{table}[H]
\centering
\resizebox{0.9 \columnwidth}{!}
{
\begin{tabular}{p{5cm} | c c | c c | c c}
\toprule
\rowcolor{gray!15} Model & \multicolumn{2}{c}{$\parle$} & \multicolumn{2}{c}{Elastic-SGD} &  \multicolumn{2}{c}{SGD}\\
 \toprule
    & Error & Time & Error & Time & Error & Time\\[0.05in]
    \rowcolor{gray!15} $\allcnn$ (full data)
    & $\bf{5.18 \pm 0.06}$ & $\bf{75}$
    & $5.76 \pm 0.07$ & $44$
    & $6.15 \pm 0.05$ & $37$\\

    $\allcnn$ ($n=3$,\ 50$\%$ data)
    & $\bf{5.89 \pm 0.01}$ & $\bf{34}$
    & $6.51 \pm 0.09$ & $36$
    & $^*7.86 \pm 0.12$ & $20$ \\

    \rowcolor{gray!15}
    $\allcnn$ ($n=6$,\ 25$\%$ data)
    & $\bf{6.08 \pm 0.05}$ & $\bf{19}$
    & $6.8 \pm 0.05$ & $20$
    & $^*10.96 \pm 0.17$ & $10$\\

\bottomrule
\end{tabular}
}
\vspace*{0.1in}
\caption{\small Splitting the dataset between replicas on CIFAR-10: Validation error (\%) at wall-clock time (min).\\ $^*$ SGD performs poorly in these cases because it only has access to a (random) subset of the training data.}
\label{tab:split_data}
\end{table}

\vspace*{-0.2in}
\begin{figure}[!htp]
\centering
    %
    \begin{subfigure}[t]{0.38\textwidth}
    \includegraphics[width=\textwidth]{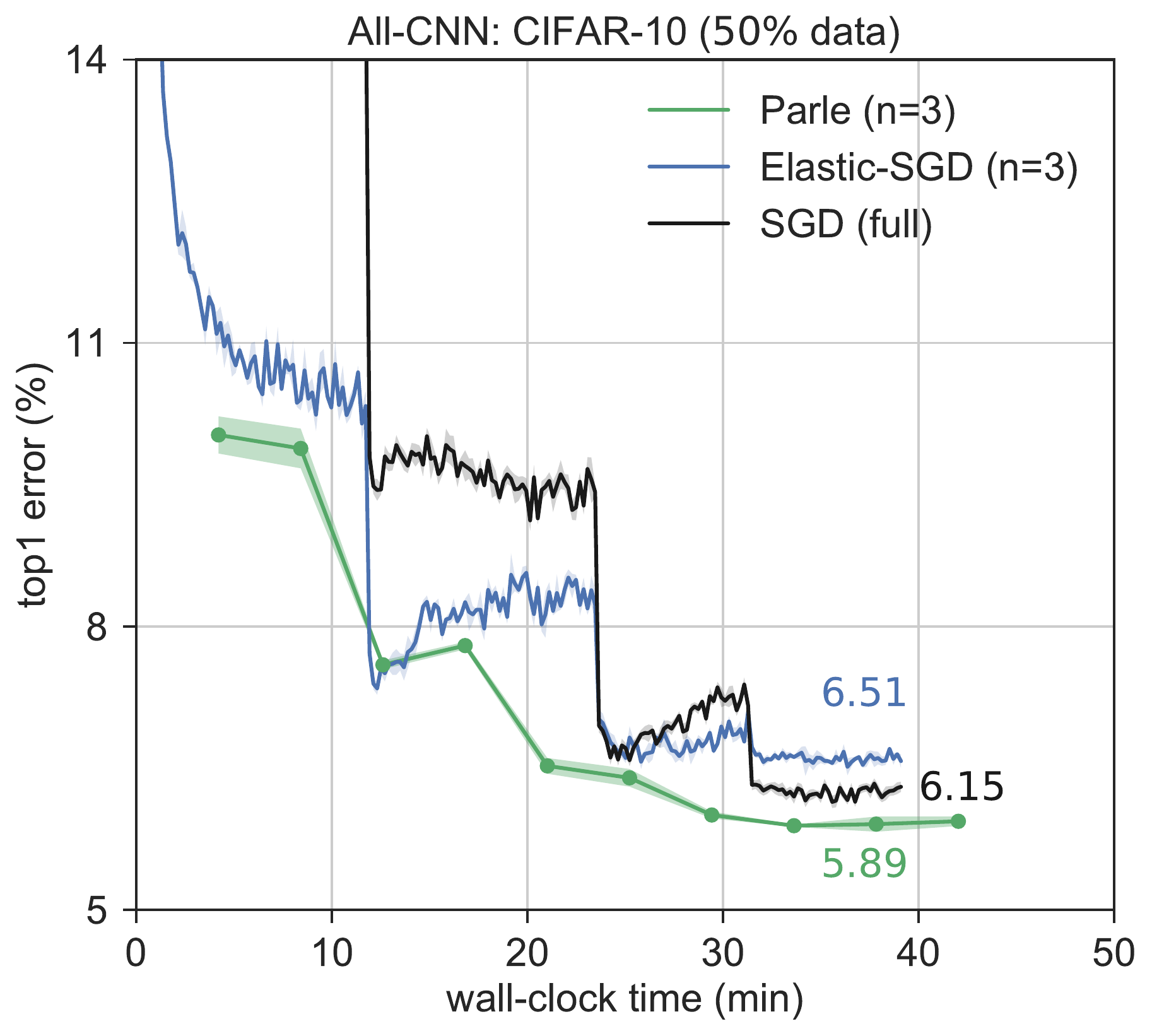}
    \caption{}
    \label{fig:allcnn_cifar10_half_valid}
    \end{subfigure}
    \hspace{0.2in}
    \begin{subfigure}[t]{0.38\textwidth}
        \includegraphics[width=\textwidth]{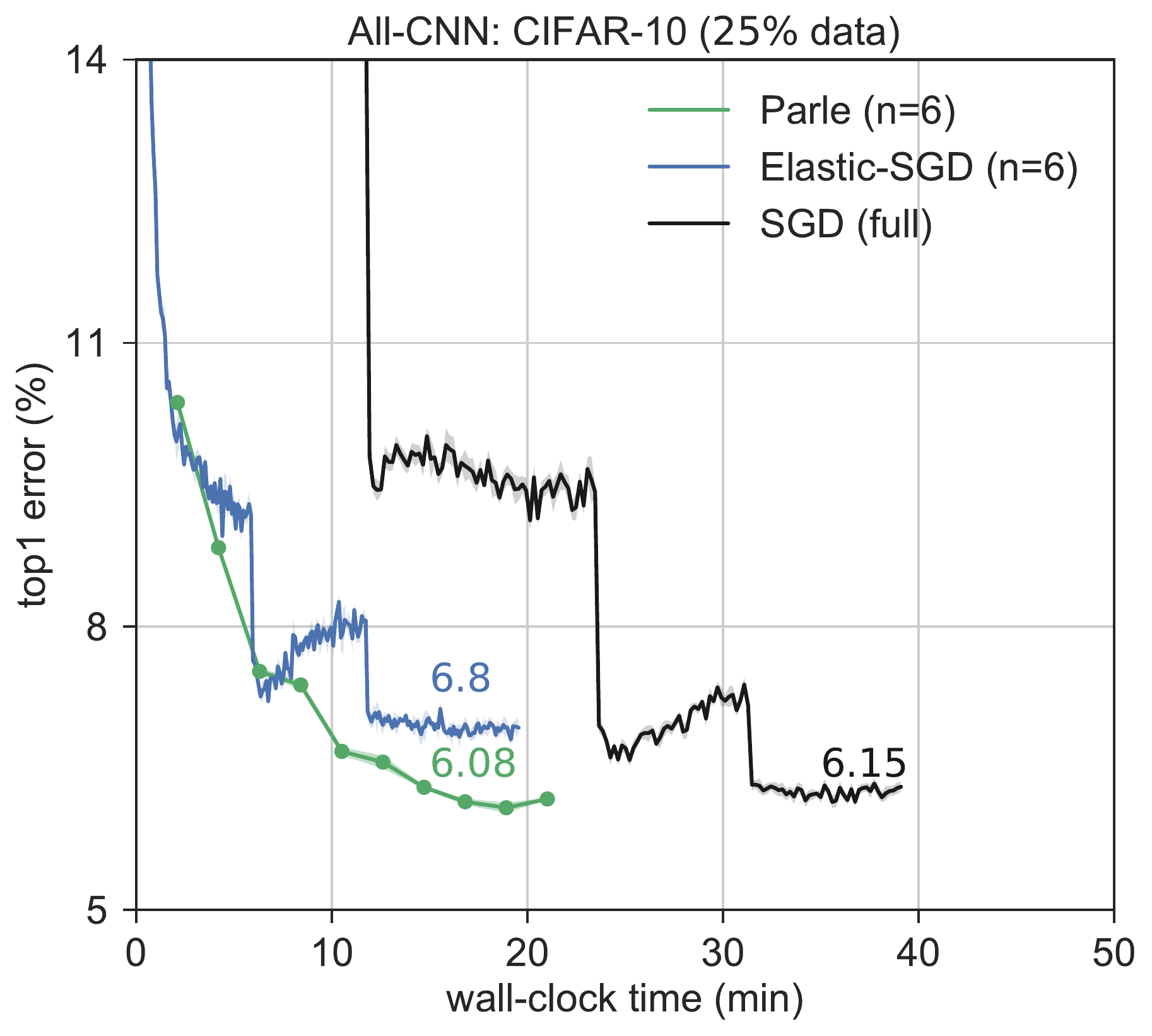}
        \caption{}
        \label{fig:allcnn_cifar10_fourth_valid}
    \end{subfigure}
\label{fig:allcnn_cifar10_frac}
\caption{\small Validation error: $\allcnn$ on CIFAR-10 for $50\%$ data (\cref{fig:allcnn_cifar10_half_valid}) and $25\%$ data (\cref{fig:allcnn_cifar10_fourth_valid}).}
\end{figure}

\section{Discussion}
This paper proposed an algorithm called $\parle$ for training deep neural networks in parallel that exploits the phenomenon of flat regions in the parameter space. $\parle$ requires infrequent communication with the parameter server and instead performs more computation on each client. It scales well to multi-GPU parallel settings and is amenable to a distributed implementation. Our experiments showed that it obtains nearly state-of-the-art performance on benchmark datasets. We obtain significantly better errors than SGD with the same architecture which shows that even with numerous regularization techniques like weight-decay, dropout and batch-normalization, there is still some performance left on the table by SGD, which $\parle$ can extract.

In the broader scope of this work, parallelization of non-convex problems like deep neural networks is fundamentally different from convex problems that have been primarily studied in distributed machine learning. Impressive large-scale distributed systems have been built for the specific purposes of deep learning but obtaining a theoretical understanding of popular heuristics used in these systems is hard. $\parle$ is a step towards developing such an understanding, for instance, the loss function used in this paper is a specific way to smooth a rugged, non-convex loss function~\citep{chaudhari2017deep}.

Another interesting offshoot of $\parle$ is that different replicas can have very different computational and communication capabilities. For instance, replicas with GPUs are more suited to run Entropy-SGD while CPU clusters and mobile devices, which can typically communicate quicker than they can compute, are more suited to run Elastic-SGD steps. Coupling these diverse platforms together in a single, interpretable loss function to train a shared model, or multiple coupled models of different sizes, is promising for both scaling up further and learning from private and sensitive data.

{
\footnotesize
\bibliographystyle{apalike}
\bibliography{chaudhari.baldassi.ea}
}

\end{document}